\documentclass{article}%
\usepackage{iclr2027_conference,times}

\usepackage{amsmath,amsfonts,bm}

\def\eqref#1{equation~\ref{#1}}
\def\1{\bm{1}}

\def\ve{{\bm{e}}}

\def\vm{{\bm{m}}}

\def\vx{{\bm{x}}}

\def\mG{{\bm{G}}}

\def\mM{{\bm{M}}}

\def\mP{{\bm{P}}}

\def\mR{{\bm{R}}}

\def\mV{{\bm{V}}}

\DeclareMathAlphabet{\mathsfit}{\encodingdefault}{\sfdefault}{m}{sl}
\SetMathAlphabet{\mathsfit}{bold}{\encodingdefault}{\sfdefault}{bx}{n}

\def\gD{{\mathcal{D}}}
\def\gE{{\mathcal{E}}}

\def\gL{{\mathcal{L}}}

\usepackage{hyperref}
\usepackage{url}
\usepackage{booktabs}
\usepackage{multirow}
\usepackage{xcolor}
\usepackage{graphicx}
\usepackage{capt-of}%
\usepackage{placeins}

\title{Enhanced Video Text Editing with Trajectory-Aligned Glyph Rendering}

\author{
Shulian Zhang$^{1}$ \quad
Xiangyu Shu$^{1}$ \quad
Wenbo Li$^{2}$ \quad
Jian Chen$^{1}$\addtocounter{footnote}{1}\thanks{Corresponding authors.} \quad
Yong Guo$^{1}$\footnotemark[2] \\
\\
$^{1}$South China University of Technology \\
$^{2}$The Chinese University of Hong Kong \\
}

\iclrfinalcopy%
\begin{document}

\maketitle
\lhead{Preprint}%

\begin{abstract}
Video text editing aims to replace or add text in a video while keeping the rest of the video unchanged, which requires the edited text to be correct in every frame and to move coherently with the scene. Despite the remarkable progress of video diffusion models, they struggle to reproduce exact stroke structures and often produce garbled or wrong characters, especially for characters with complex strokes. To address this, we propose a \textbf{trajectory-aligned glyph rendering reference} that provides explicit per-frame glyph guidance following the position and perspective of the text, and a \textbf{depth-normalized recognizer feature supervision} that supervises the generated text on multi-depth features of a frozen text recognizer with per-depth normalized errors, targeting stroke errors overlooked by the diffusion loss. We further build \textbf{VTEdit}, a benchmark of 288 real-scene clips with 440 annotated text trajectories covering text replacement and text addition, which will be publicly released to facilitate future research. Experiments on VTEdit show that our method outperforms image text editing methods, video editing methods, and commercial models in text accuracy and background preservation, achieving a sentence accuracy of 0.9408, and receives the highest preference in a user study.
\end{abstract}

\section{Introduction} \label{sec:intro}

Video text editing replaces the existing text in a video, or adds new text to it, while leaving the rest of the video unchanged. It serves applications such as advertisement design and film post-production. Compared with editing a single image, editing text in a video is harder in two respects. First, the text must be correct in every frame. A single wrong or missing stroke is easy for a reader to notice, and an error in any frame spoils the whole video. Second, the text must blend naturally into the scene and stay attached to the surface it is written on, following its position and perspective as the camera or the object moves, without flickering between frames.

However, existing video editing methods still struggle to edit text accurately and coherently. General video editing models~\citep{jiang2025vace,bian2025videopainter,wei2026univideo,lin2026kiwiedit} maintain temporal coherence, but they are built for scene- and object-level edits rather than fine-grained text rendering. They offer no glyph-level control over the edited region, and the text they generate is often garbled, especially for characters with complex strokes. Image text editing methods~\citep{tuo2024anytext2,lan2025fluxtext} achieve accurate glyphs, but they are designed for single images and do not model the relation between frames. Only a few methods are dedicated to video text editing. STRIVE~\citep{g2021strive} replaces the text in a single frontalized reference frame and then propagates it to the other frames, so errors in any stage of this multi-stage pipeline carry over to the final video. SteerVTE~\citep{zeng2026steervte} edits videos end to end, but supports only English text. Progress is further limited by data for both training and evaluation. For training, paired videos that differ only in their text can hardly be collected in real scenes. For evaluation, to the best of our knowledge, no public benchmark of real-scene videos exists for video text editing.

Accurate text rendering remains a fundamental challenge for video generation models. A character is defined by its exact strokes, and a single wrong stroke turns it into a different or nonexistent character. Video generation models, however, are built to produce content that looks plausible rather than exact, and they struggle to reproduce the precise stroke structure of characters, especially for Chinese, with thousands of characters and many complex strokes. The difficulty is compounded in videos, where the glyphs must further follow the position and perspective of the text as it moves. Instead of expecting the model to learn the exact strokes of every character, our key idea is to provide them explicitly. When the model is shown what the target glyphs look like and where they should appear in each frame, it can follow the provided strokes instead of generating them on its own, and blend them into the scene by adapting their color, texture, and lighting to the surroundings. The provided glyphs thus serve as a direct reference for the exact strokes, while the generative model focuses on what it does best, namely producing realistic appearance. The same bias toward plausibility persists in training. The diffusion loss measures how closely the generated video matches the target as a whole, to which a single stroke contributes little, so a wrong stroke is barely penalized. We therefore complement the diffusion loss with a text-aware supervision, which evaluates the generated text with prior knowledge of how text is read rather than by its overall appearance alone.

To this end, we formulate video text editing as text-conditioned video inpainting along a text trajectory, i.e., a sequence of per-frame text boxes: the content inside the boxes is regenerated, and the content outside is preserved. Since the original text inside the boxes is hidden from the model, a real video can serve as its own ground truth, which enables training on real videos without paired data. Under this formulation, we realize the two ideas above with two components. First, the \textit{trajectory-aligned glyph rendering reference} renders the target text and warps it into the text box of every frame, so that the provided glyphs follow the position and perspective of the text throughout the video. The rendered video is fed to the model as a spatially aligned condition, together with a glyph-aware text encoder that specifies the identity of each character. Second, we supervise the generated text with the intermediate features of a frozen text recognizer at multiple depths, from shallow features that retain stroke details (Figure~\ref{fig:featdepth}) to deep features more related to character identity. We observe that the activations at different depths lie on different scales (Figure~\ref{fig:featdist}), so that summing their errors directly lets the depth with larger activations dominate. Our \textit{depth-normalized recognizer feature supervision} therefore normalizes the error at each depth by the magnitude of the ground-truth features at that depth, balancing the depths without per-depth weights. For training and evaluation, we collect real-scene videos annotated with text trajectories, from which we build VTEdit, a benchmark for text replacement and text addition.

Overall, we make three key contributions: 1) We propose a \textbf{\textit{trajectory-aligned glyph rendering reference}}, which renders the target text into the text box of every frame and conditions video generation on it, providing explicit glyph guidance that follows the position and perspective of the text. 2) We introduce \textbf{\textit{depth-normalized recognizer feature supervision}}, which supervises the generated text on the decoded frames with multi-depth features of a frozen text recognizer, normalized per depth to balance their contributions. 3) We build \textbf{\textit{VTEdit}}, a real-scene video text editing benchmark with 440 annotated text trajectories, to be publicly released. On VTEdit, our method achieves the best text accuracy and background preservation, improving Sen.ACC from 0.5980 to 0.9408 over the strongest video editing model, Seedance, and receiving 64.3\% of the user votes against its 30.8\%.

\section{Related Work}
\label{sec:related}

\paragraph{Video Editing and Visual Text Editing.}
Diffusion-based video editing has evolved from adapting image diffusion models, through inter-frame feature propagation~\citep{geyer2024tokenflow} or per-video tuning~\citep{wu2023tuneavideo}, to building on video foundation models~\citep{kong2024hunyuanvideo,wanteam2025wan}. Recent frameworks support mask-conditioned editing and inpainting~\citep{jiang2025vace,bian2025videopainter} as well as instruction-guided editing~\citep{wei2026univideo,lin2026kiwiedit}, but they target scene- and object-level edits and offer no glyph-level control. Visual text generation and editing has instead been studied mainly for images, where diffusion models are conditioned on character-level layouts, glyph images, or text positions to render accurate text~\citep{chen2023textdiffuser,chen2023diffute,ma2023glyphdraw,tuo2024anytext,tuo2024anytext2,lan2025fluxtext}, and recent methods further address instruction-guided editing~\citep{ma2026umtext}, stylized text in graphic design~\citep{zhao2025utdesign}, and bilingual Chinese and English text~\citep{liu2026innotext}. When applied to videos frame by frame, these methods do not model the relation between frames, so the edited text may vary across frames. For video, STRIVE~\citep{g2021strive} replaces the text in a frontalized reference frame and propagates it to the other frames, so errors in any stage of this multi-stage pipeline carry over to the final video. SteerVTE~\citep{zeng2026steervte} injects style features and line- and character-level glyph images into a frozen video diffusion transformer through cross-attention, and supports only English text.

\paragraph{Recognizer-based Text Supervision.}
Since the diffusion loss is barely affected by errors in individual strokes, several methods additionally supervise the generated text in pixel space with pretrained text models. OCR-VQGAN~\citep{DBLP:conf/wacv/RodriguezVLPR23} trains an image autoencoder with a perceptual loss on multi-layer features of a text detector, normalized along the channel dimension. AnyText~\citep{tuo2024anytext} compares the features of a text recognizer before its final fully connected layer between the generated and ground-truth text regions, i.e., at a single depth. JoyType~\citep{DBLP:journals/corr/abs-2409-17524} and CharGen~\citep{DBLP:journals/corr/abs-2412-17225} extend this comparison to multiple layers, using the early convolutional layers of an OCR model and the multi-scale features of an OCR destylization model, respectively, and SteerVTE~\citep{zeng2026steervte} brings multi-layer recognizer features, together with a CTC loss, to video text editing. Other methods fine-tune generators via reinforcement learning with rewards computed from recognition results~\citep{DBLP:conf/nips/LiuLLLLWWZO25}, which evaluate the recognized text rather than intermediate features. In the multi-layer losses of JoyType, CharGen, and SteerVTE, the error of each layer is normalized only by its feature size before summation, so layers with larger activations can dominate, and the balance among layers depends on which layers are chosen. In contrast, we supervise the generated text at shallow, middle, and deep depths of a frozen text recognizer and normalize the error at each depth by the magnitude of the ground-truth features, placing all depths on a comparable scale without per-depth weights.

\section{Enhanced Video Text Editing}
\label{sec:method}

\subsection{Problem definition}
\label{sec:problem}
We formulate video text editing as text-conditioned video inpainting along a text trajectory. Given a source video $\mV$ with $F$ frames, a text trajectory $\mP=\{\mP_i\}_{i=1}^{F}$ where $\mP_i$ is the text box in frame $i$, a target text $y$, and a scene prompt $c$, the model generates an edited video $\hat{\mV}$ by regenerating the content inside the boxes, marked by the mask $\mM=\{\mM_i\}_{i=1}^{F}$ ($\mM_i$ is $1$ inside $\mP_i$ and $0$ elsewhere). The generated text should follow the box in every frame, blend into the scene, and remain consistent across frames, and the content outside the boxes should remain unchanged. Since the original text inside the boxes is hidden from the model, a real video can serve as its own ground truth: during training, $y$ is set to the original text of the video, while at inference $y$ is set to the desired text.

\subsection{Trajectory-Aligned Glyph Rendering Reference}
\label{sec:method_3_1}
\begin{figure}[t]
    \centering
    \includegraphics[width=\linewidth]{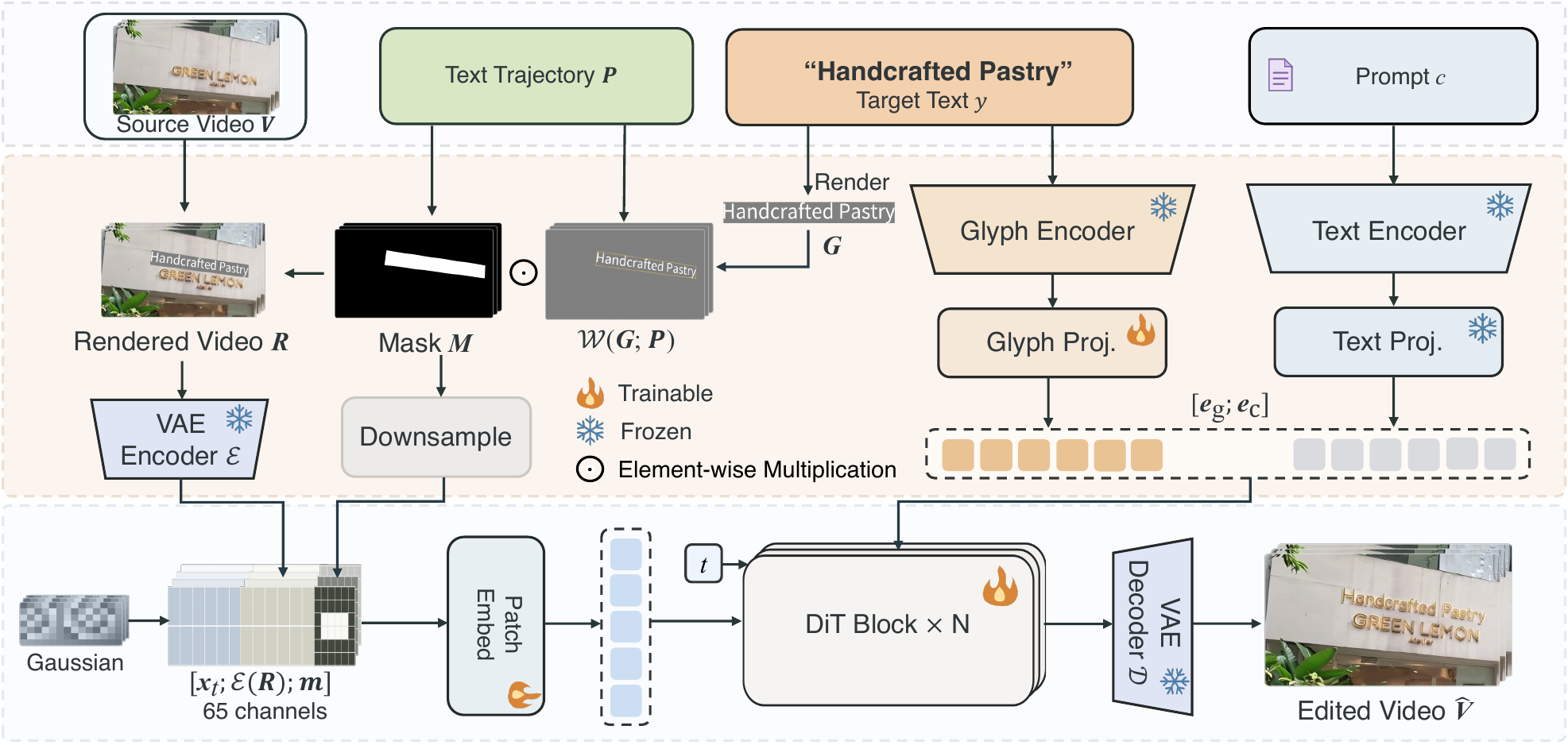}
    \caption{Overview of our framework with the trajectory-aligned glyph rendering reference. The target text $y$ is rendered into a glyph image $\mG$, which is warped into the text box of every frame along the trajectory $\mP$. The warped glyphs $\mathcal{W}(\mG;\mP)$ replace the content of the source video $\mV$ inside the mask $\mM$, forming the rendered video $\mR$ (Eq.~\ref{eq:reference}). The latent $\gE(\mR)$ and the downsampled mask $\vm$ are concatenated with the noisy latent $\vx_t$ along the channel dimension, and the glyph embedding $\ve_g$ is concatenated with the prompt tokens $\ve_c$ along the sequence dimension, as the inputs of the diffusion transformer. The VAE decoder $\gD$ decodes the generated latent into the edited video $\hat{\mV}$.}
    \label{fig:pipeline}
\end{figure}

Video generation models often fail to reproduce the precise strokes of complex characters and produce garbled or wrong text. Instead of requiring the model to generate the strokes on its own, we render the target glyphs and provide them as a reference, which the model follows and blends into the scene. Since the text moves and changes in perspective across frames, we render $y$ along the text trajectory, so that the glyphs follow the position and perspective of the text throughout the video. We further encode $y$ with a glyph-aware text encoder, Glyph-ByT5~\citep{liu2024glyph}, which specifies the identity of each character. Figure~\ref{fig:pipeline} shows an overview.

Concretely, we first draw $y$ in white on a gray canvas with a fixed font, vertically for vertical text, and crop tightly around the glyphs to obtain a glyph image $\mG$, which is shared across all frames. For each frame $i$, we warp $\mG$ onto the frame with the perspective transform that maps the four corners of $\mG$ to the four vertices of $\mP_i$, denoted $\mathcal{W}(\mG;\mP_i)$. To reduce aliasing and blurring of the strokes, $\mG$ is drawn at a large font size so that the warp mostly shrinks it, and the warp is performed on a canvas with $4\times$ the frame width and height and then downsampled to the frame resolution by area averaging. The warped glyphs replace the content of the text box in the source frame:
\begin{equation}
\mR_i=(1-\mM_i)\odot\mV_i+\mM_i\odot\mathcal{W}(\mG;\mP_i),
\label{eq:reference}
\end{equation}
where $\mV_i$ and $\mM_i$ are the $i$-th frames of $\mV$ and $\mM$, and $\odot$ denotes element-wise multiplication. Frames without text have $\mM_i=\mathbf{0}$ and remain unchanged, and $\{\mR_i\}_{i=1}^{F}$ form the rendered video $\mR$. Since the training target is the original text in the real video, whose appearance generally differ from those of the rendered glyphs, the reference guides the structure of the glyphs rather than their appearance. Besides $\mR$, the glyph-aware text encoder encodes $y$ into a glyph embedding $\ve_g$.

We then feed this reference into a latent video diffusion transformer. A VAE encoder $\gE$ maps the rendered video $\mR$ to a latent $\gE(\mR)$ of the same size as the noisy latent $\vx_t$, and the mask $\mM$ is downsampled to the same latent resolution to obtain $\vm$. We concatenate $\vx_t$, $\gE(\mR)$, and $\vm$ along the channel dimension, so that the rendered glyphs are spatially aligned with the latent to be generated. The glyph embedding $\ve_g$ is concatenated with the prompt tokens $\ve_c$. The transformer $v_\theta$ then predicts the flow-matching velocity~\citep{lipman2023flowmatching}:
\begin{equation}
v_\theta\big([\vx_t;\gE(\mR);\vm],\,t,\,[\ve_g;\ve_c]\big),
\label{eq:condition}
\end{equation}
where $t\in[0,1]$ is the timestep, and $[\cdot\,;\cdot]$ denotes concatenation along the channel dimension for latents and the sequence dimension for tokens. This reference improves text accuracy (Section~\ref{ssec:ablation}).

\subsection{Depth-Normalized Recognizer Feature Supervision}
\label{sec:ocr_loss}

\begin{figure}[t]
    \centering
    \includegraphics[width=\linewidth]{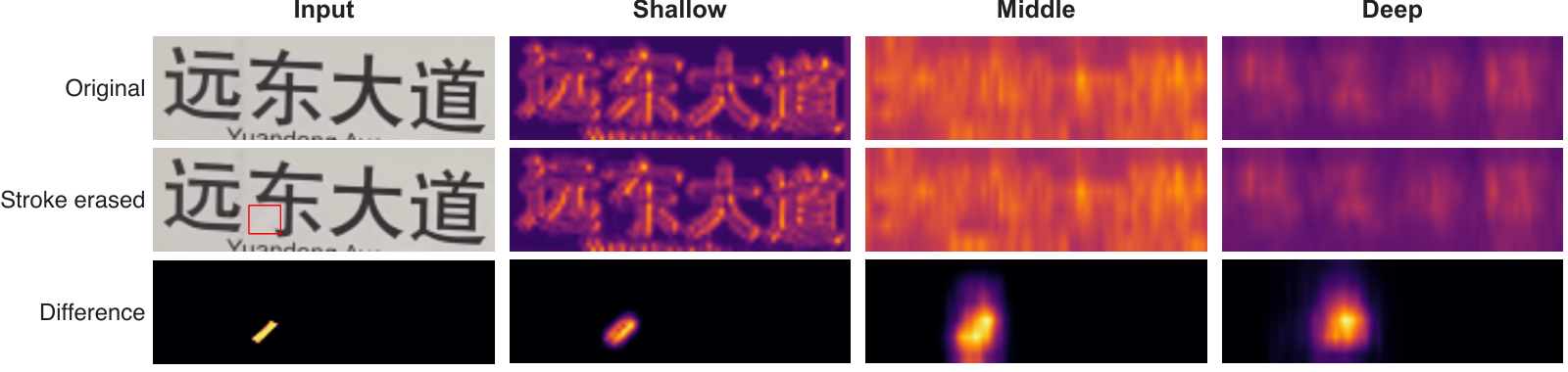}
    \caption{Visualization of perceptual features at the three supervised depths of the text recognizer PP-OCRv3. For clarity, we manually erase a stroke (red box) to observe the differences. At the shallow depth, the feature difference is sharp and localized to the missing stroke; at the middle and deep depths, it becomes diffuse and covers a larger part of the character.}
    \label{fig:featdepth}
\end{figure}

\begin{figure}[t]
    \centering
    \includegraphics[width=0.49\linewidth]{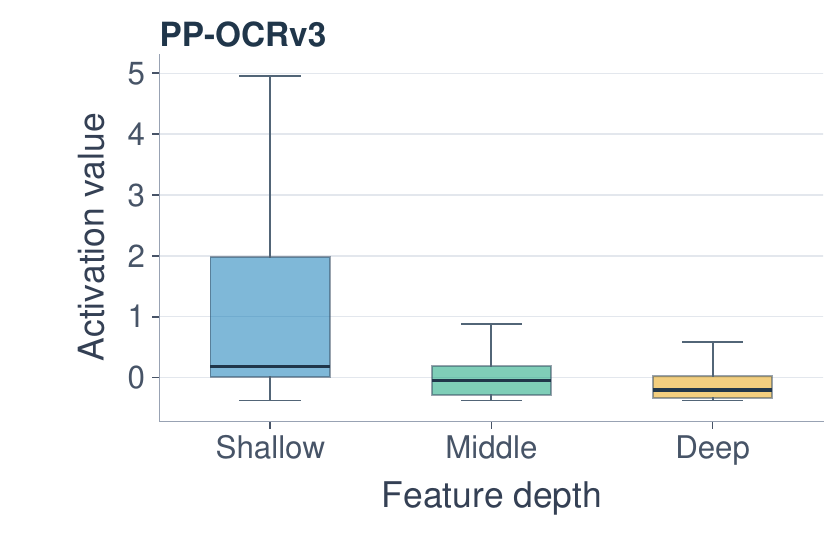}
    \hfill
    \includegraphics[width=0.49\linewidth]{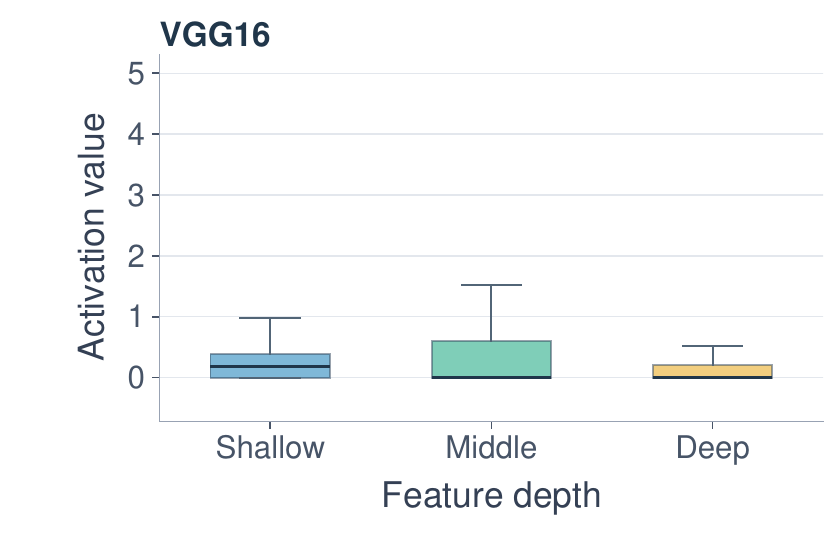}
    \caption{Distribution of the activation values at the three supervised depths, measured on the text regions. Left: in the text recognizer, the three depths lie on very different scales, with the shallow depth reaching about 5 while the middle and deep depths stay below 1. Right: in VGG, the three depths lie on comparable scales.}
    \label{fig:featdist}
\end{figure}

\begin{figure}[t]
    \centering
    \includegraphics[width=\linewidth]{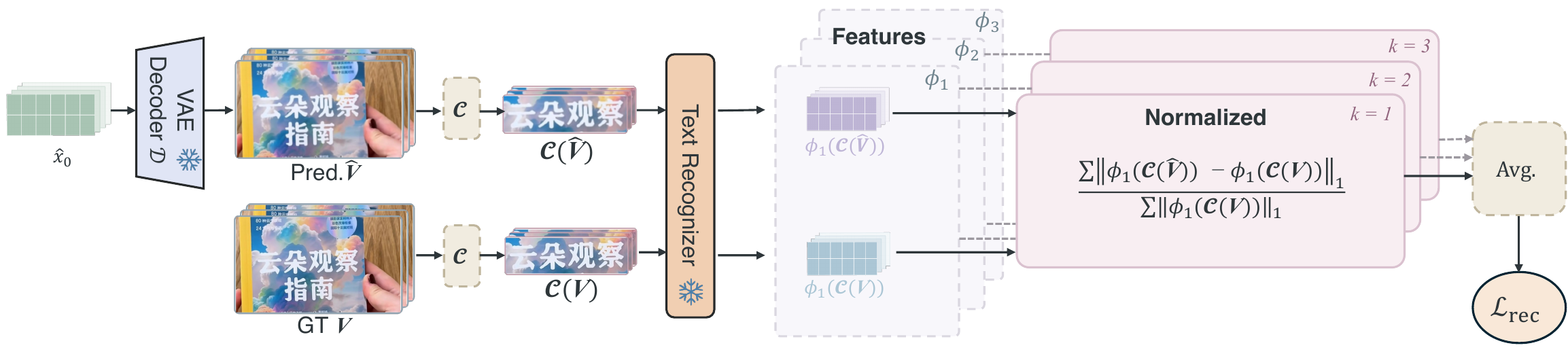}
    \par\vspace{-5pt}
    \caption{Overview of the depth-normalized recognizer feature supervision. During training, the frozen VAE decoder $\gD$ decodes the estimated clean latent $\hat{\vx}_0$ into the predicted video $\hat{\mV}$. The text regions of $\hat{\mV}$ and the ground truth $\mV$ are cropped by $\mathcal{C}$ and passed through a frozen text recognizer, which extracts the features $\phi_1$, $\phi_2$, and $\phi_3$ at the shallow, middle, and deep depths. The error at each depth is normalized by the magnitude of the ground-truth features, and the three normalized errors are averaged into $\gL_{\text{rec}}$ (Eq.~\ref{eq:rec}), which is added to the flow-matching loss.
    }
    \label{fig:rec_loss}
\end{figure}

The flow-matching loss measures how closely the generated video matches the target as a whole, to which a single wrong stroke contributes little. We therefore additionally supervise the text in the decoded frames with a frozen text recognizer. Since a recognizer may still recognize a character with a missing stroke~\citep{DBLP:journals/corr/abs-2412-17225}, we compare its intermediate features rather than its final output. To see how these features respond to a stroke error, we erase a single stroke and inspect the feature difference at each depth (Figure~\ref{fig:featdepth}): at the shallow depth the difference is sharp and localized to the missing stroke, while at the middle and deep depths it becomes diffuse and covers a larger part of the character. This suggests that shallow features retain individual strokes, whereas deeper features, being closer to the recognition head, are more related to character identity, consistent with the observation that deeper CNN features are more class-specific~\citep{DBLP:conf/eccv/ZeilerF14}. To capture both stroke details and character identity, we supervise at three depths, shallow, middle, and deep. However, we observe that the three depths of the recognizer lie on very different scales (Figure~\ref{fig:featdist}), with the shallow depth spanning a much wider range than the middle and deep depths, whereas those of a natural-image network such as VGG~\citep{DBLP:journals/corr/SimonyanZ14a} lie on comparable scales. Since the error at each depth grows with the scale of its activations, summing the three errors directly lets the depth with the largest activations dominate. We therefore propose depth-normalized recognizer feature supervision (Figure~\ref{fig:rec_loss}), which normalizes the error at each depth by the magnitude of the ground-truth features at that depth, balancing the three depths without per-depth weights.

Since the recognizer operates on images, we apply the supervision to frames decoded from the prediction of the model. During training, the noisy latent is constructed as $\vx_t=(1-t)\vx_0+t\bm{\epsilon}$, where $\vx_0=\gE(\mV)$ is the latent of the source video and $\bm{\epsilon}\sim\mathcal{N}(\mathbf{0},\mathbf{I})$ is Gaussian noise. From the velocity predicted in Eq.~\ref{eq:condition}, we estimate the clean latent as $\hat{\vx}_0=\vx_t-t\,v_\theta(\cdot)$ and decode it with the VAE decoder $\gD$ into the predicted frames $\hat{\mV}=\gD(\hat{\vx}_0)$. For each frame $i$ that contains the text, an operation $\mathcal{C}_i(\cdot)$ crops the bounding rectangle of the text box $\mP_i$, rotates vertical text to horizontal, and resizes the crop to the input size of the recognizer. The crops of $\hat{\mV}$ and $\mV$ are passed through the recognizer backbone, from which we take the features $\phi_k$ at the $k$-th of the $K=3$ depths (shallow, middle, and deep). The supervision is defined as
\begin{equation}
\gL_{\text{rec}}=\frac{1}{K}\sum_{k=1}^{K}\frac{\sum_{i\in\mathcal{S}}\big\|\phi_k\big(\mathcal{C}_i(\hat{\mV})\big)-\phi_k\big(\mathcal{C}_i(\mV)\big)\big\|_1}{\sum_{i\in\mathcal{S}}\big\|\phi_k\big(\mathcal{C}_i(\mV)\big)\big\|_1},
\label{eq:rec}
\end{equation}
where $\mathcal{S}$ is the set of decoded frames that contain the text and $\|\cdot\|_1$ denotes the sum of absolute values over all feature elements. The recognizer and the VAE decoder are frozen, and the gradient of $\gL_{\text{rec}}$ is propagated through $\gD$ to $v_\theta$. Both supervising multiple depths and normalizing each depth improve text accuracy in our ablation studies (Section~\ref{ssec:ablation}).

\subsection{Training and Inference Method}
\label{sec:training}
We build our model on the 480p text-to-video model of HunyuanVideo~1.5~\citep{hunyuanvideo2025}, with Glyph-ByT5~\citep{liu2024glyph} as the glyph-aware text encoder and Qwen2.5-VL~\citep{Qwen2.5-VL} as the prompt encoder. The base model and both text encoders are frozen, and we train LoRA~\citep{hu2022lora} with rank 32 on all transformer blocks, together with the projector of the glyph embedding $\ve_g$ and the patch embedding layer. The flow-matching loss is computed over the whole latent,
\begin{equation}
\gL_{\text{FM}}=\mathbb{E}_{t,\bm{\epsilon}}\big\|v_\theta(\cdot)-(\bm{\epsilon}-\vx_0)\big\|_2^2,
\label{eq:fm}
\end{equation}
where $v_\theta(\cdot)$ is the prediction in Eq.~\ref{eq:condition}, and the model is trained with
\begin{equation}
\gL=\gL_{\text{FM}}+\gL_{\text{rec}}.
\label{eq:total}
\end{equation}
For $\gL_{\text{rec}}$, we use the backbone of the PP-OCRv3 recognizer~\citep{li2022pp} and take the outputs of its 1st, 6th, and 10th blocks as the features $\phi_1$, $\phi_2$, and $\phi_3$ at the shallow, middle, and deep depths. To save memory, only the first 9 frames of $\hat{\mV}$ are decoded by $\gD$ for $\gL_{\text{rec}}$. The training data consist of 1{,}050 real-scene clips with 1{,}430 text trajectories, collected and annotated with the same pipeline as VTEdit (Section~\ref{sec:benchmark}) and disjoint from it. Each training sample is a 33-frame clip spatially resized to $480\times832$. We use the Muon optimizer~\citep{jordan2024muon} with a learning rate of $1\times10^{-4}$ and a batch size of 8, and train for 30{,}000 steps in total on 8 Ascend 910B NPUs.

At inference, each clip of up to 49 frames is resized to $480\times832$ and edited as a whole, and the model generates the edited video from Gaussian noise with 50 sampling steps. When a video contains multiple text trajectories, we edit them sequentially, taking the output of each edit as the source video of the next, so that all edits are kept in one video.

\section{VTEdit: Video Text Editing Benchmark}
\label{sec:benchmark}
\begin{figure}[t]
\centering
\includegraphics[width=\linewidth]{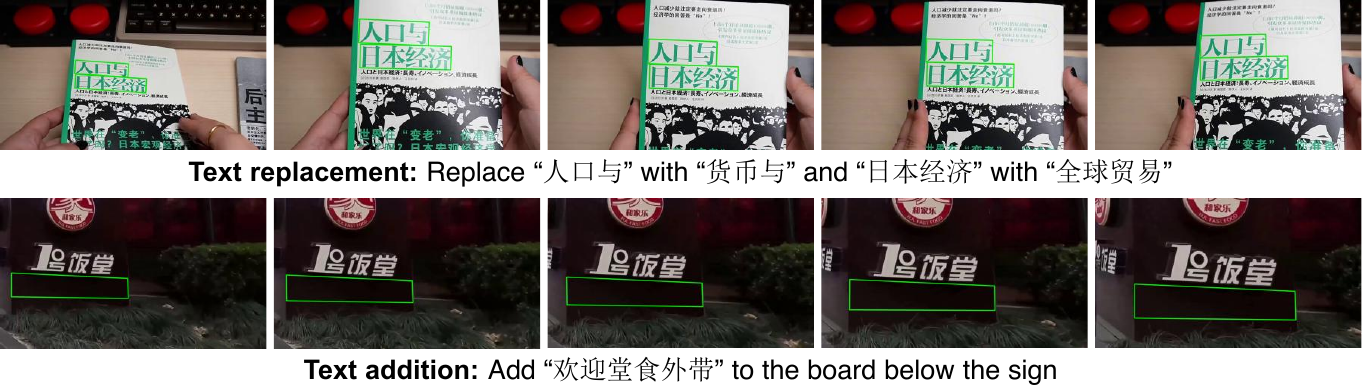}
\caption{Examples of text replacement and text addition in VTEdit. Each row shows five frames sampled from a clip, with the annotated text boxes in green, and the text below describes the edit.}
\label{fig:benchmark}
\end{figure}

While scene text editing has been widely benchmarked on static images, to the best of our knowledge, no public benchmark yet exists for video text editing on real-scene videos. To fill this gap, we build a Video Text Editing benchmark (VTEdit) from 4K real-scene videos collected from public social-media platforms. We select clips containing scene text, crop a region from each clip, and resize it to $720\times1280$. Each text instance is annotated as a trajectory of quadrilateral boxes over frames: the text is detected with PP-OCRv6~\citep{ppocrv62026} and associated across frames. Each clip is further captioned with Qwen2.5-VL~\citep{Qwen2.5-VL}. The benchmark contains 288 clips with 440 trajectories, disjoint from the training set and each trimmed to at most 49 frames, following SteerVTE~\citep{zeng2026steervte}. It covers two tasks, shown in Figure~\ref{fig:benchmark}: text replacement (398 trajectories), where the original text is replaced by a target of similar length, and text addition (42 trajectories in 30 clips), where new text is written onto a text-free surface, whose box is manually drawn in the first frame and propagated to the other frames by a tracking algorithm. The target texts of both tasks are generated by GPT-5~\citep{DBLP:journals/corr/abs-2601-03267}. Chinese text appears in 358 trajectories and English text in 82, and the Chinese text covers 740 distinct characters.

\section{Experiments} \label{sec:exp}
In this section, we evaluate our method on the VTEdit benchmark, with the metrics and compared methods described in Section~\ref{ssec:implementation}. In Section~\ref{ssec:sota}, our method achieves the highest text accuracy, background preservation, and user preference among image text editing methods, mask-based and instruction-guided video editing methods, and commercial models. The ablation studies in Section~\ref{ssec:ablation} show that the glyph rendering reference, the supervision at multiple depths of the text recognizer, and the depth-wise normalization each improve text accuracy.

\subsection{Experimental details}
\label{ssec:implementation}
\noindent\textbf{Evaluation and Metrics.}
We evaluate text accuracy, temporal consistency, and background preservation, and further conduct a user study. All metrics are computed after resizing the outputs of all methods and the source videos to $480\times832$. Text accuracy is measured by sentence accuracy (Sen.ACC) and normalized edit distance (NED)~\citep{tuo2024anytext}, using PP-OCRv6~\citep{ppocrv62026}, a recognizer different from the one used in training. Both are computed on the text box in each frame and averaged over all frames. Temporal consistency is measured by the warping error $E_{warp}$~\citep{E_warp_2018_ECCV} with RAFT~\citep{teed2020raft} optical flow. Background preservation is measured by PSNR and SSIM~\citep{ssim} between the edited and source videos over the region outside the text boxes, denoted PSNR$_{bg}$ and SSIM$_{bg}$. In the user study, six participants view the anonymized results of all methods on 100 cases from VTEdit and select the best one in each case considering text accuracy, natural blending, temporal consistency, and background preservation. The user preference of a method is the percentage of votes it receives.

\noindent\textbf{Compared Methods.}
We compare with three groups of methods: the image text editing method FLUX-Text~\citep{lan2025fluxtext}, applied to each frame independently; mask-based video editing methods, which regenerate the masked region of a video, including VACE~\citep{jiang2025vace} and VideoPainter~\citep{bian2025videopainter}; and instruction-guided video editing methods, which edit a video according to a text instruction, including UniVideo~\citep{wei2026univideo}, Kiwi-Edit~\citep{lin2026kiwiedit}, and the closed-source commercial models Kling O3~\citep{klingteam2025klingomni} and Seedance 2.0 Fast~\citep{seedance2026seedance}, denoted Seedance for brevity. The inputs and settings used for each compared method are detailed in Appendix~\ref{app:baselines}.

\subsection{Comparison with State-of-the-Art}
\label{ssec:sota}

\textbf{Quantitative Results.}
Table~\ref{tab:sota} reports the quantitative comparison on VTEdit. Our method achieves the best text accuracy, background preservation, and user preference, while maintaining temporal consistency comparable to the video editing methods. Among the compared methods, the image text editing method FLUX-Text achieves the highest text accuracy, but editing each frame independently results in the largest warping error. The video editing methods achieve lower warping errors, but their text accuracy is considerably lower, with the best Sen.ACC of 0.5980 obtained by Seedance. Compared with FLUX-Text, our method improves Sen.ACC from 0.8393 to 0.9408 and NED from 0.9307 to 0.9862, and reduces the warping error from 4.1516 to 1.5433, close to the lowest value of 1.4862 obtained by Seedance. For background preservation, FLUX-Text and the mask-based video editing methods, which take the editing region as input, achieve PSNR$_{bg}$ above 26\,dB, whereas the instruction-guided methods, which receive no editing region, achieve about 20\,dB. Our method achieves the highest PSNR$_{bg}$ of 36.11\,dB and SSIM$_{bg}$ of 0.9736. In the user study, which considers natural blending, our method receives 64.3\% of the votes, followed by Seedance with 30.8\%.

\begin{table}[t]
\centering
\caption{Quantitative comparison with state-of-the-art methods. The best and second-best
results are highlighted in \textbf{bold} and \underline{underlined}, respectively.
$^\dagger$ denotes a closed-source commercial model. Our method achieves the best text accuracy, background preservation, and user preference.}
\label{tab:sota}
\small
\begin{tabular*}{\linewidth}{@{\extracolsep{\fill}} l cc c cc c}
\toprule
\multirow{2}{*}{Method}
  & \multicolumn{2}{c}{Text Accuracy}
  & Temporal
  & \multicolumn{2}{c}{Background Preservation}
  & User Study \\
\cmidrule(lr){2-3}\cmidrule(lr){4-4}\cmidrule(lr){5-6}\cmidrule(lr){7-7}
  & Sen.ACC$\uparrow$ & NED$\uparrow$
  & $E_{warp}$$\downarrow$
  & PSNR$_{bg}$$\uparrow$ & SSIM$_{bg}$$\uparrow$
  & Preference (\%)$\uparrow$ \\
\midrule
FLUX-Text          & \underline{0.8393} & \underline{0.9307} & 4.1516 & 34.55 & \underline{0.9678} & 0.3 \\
VACE               & 0.1532 & 0.2991 & 1.5593 & \underline{34.67} & 0.9669 & 1.2 \\
VideoPainter       & 0.0395 & 0.0945 & 2.0111 & 26.72 & 0.9423 & 0.3 \\
UniVideo           & 0.0002 & 0.0710 & 1.7399 & 20.04 & 0.6689 & 0.2 \\
Kiwi-Edit          & 0.0000 & 0.0862 & 1.5849 & 20.20 & 0.7143 & 0.0 \\
Kling O3$^\dagger$ & 0.1888 & 0.4577 & 2.4885 & 20.03 & 0.6644 & 2.8 \\
Seedance$^\dagger$ & 0.5980 & 0.7320 & \textbf{1.4862} & 20.36 & 0.6835 & \underline{30.8} \\
\textbf{Ours} & \textbf{0.9408} & \textbf{0.9862} & \underline{1.5433} & \textbf{36.11} & \textbf{0.9736} & \textbf{64.3} \\
\bottomrule
\end{tabular*}
\end{table}

\begin{figure}[t]
    \centering

    \includegraphics[width=\linewidth]{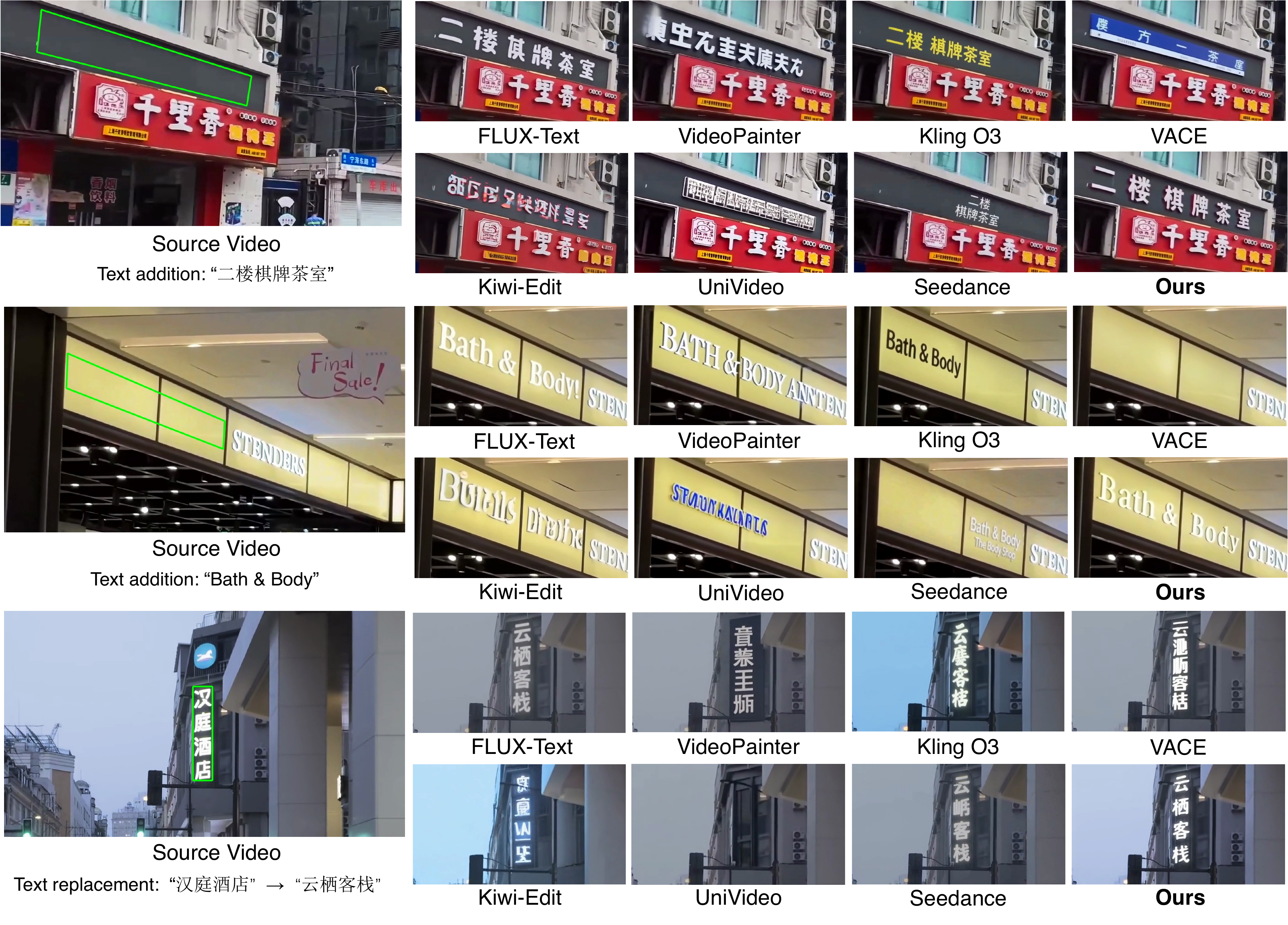}

    \caption{\textbf{Qualitative comparison of video text editing.} The top two examples show text addition, and the bottom example shows text replacement. For each example, the source frame is shown on the left, and the results of all methods are shown on the right. Green boxes in the source frames indicate the editing regions. Our results show legible text with scene-compatible layout
and appearance.}
    \label{fig:main_figs1}
    \vspace{-10pt}
\end{figure}

\textbf{Qualitative Results.}
Figure~\ref{fig:main_figs1} shows qualitative comparisons on two text addition examples and one text replacement example. VideoPainter, VACE, UniVideo, and Kiwi-Edit mostly generate wrong or garbled characters. In addition, VACE adds no text in the second example, VideoPainter also alters the adjacent text ``STENDERS'' outside the editing region, and UniVideo removes the original text in the third example without writing the target. The commercial models generate more legible text but do not consistently match the target or the scene. Kling O3 renders the correct text in the two addition examples, but the text does not blend naturally into the scene, and it generates wrong characters in the replacement example. Seedance renders the correct text in the addition examples at a scale much smaller than the editing region, adds the unrequested phrase ``The Body Shop'' in the second example, and generates a wrong character in the replacement example. FLUX-Text renders mostly correct text in individual frames, but since it edits each frame independently, its text varies across frames (Figure~\ref{fig:temp_consistency}). Our method renders the correct text in all three examples, follows the position and perspective of the editing region, and matches the appearance of the surrounding text, such as the white serif letters of ``STENDERS'' in the second example and the illuminated vertical sign in the third example. Additional qualitative comparisons focusing on text accuracy, temporal consistency, and background preservation are provided in Figures~\ref{fig:text_acc}--\ref{fig:bg_preservation} of Appendix~\ref{app:qualitative}.

\subsection{Ablation Studies}
\label{ssec:ablation}
We ablate both components of our method on VTEdit by removing the glyph rendering reference, removing the recognizer feature supervision, replacing its features with those of a single depth or of VGG, and removing its depth-wise normalization. The results show that the glyph rendering reference, the supervision at multiple depths of the recognizer, and the depth-wise normalization each improve text accuracy, with the reference bringing the largest gain.

\noindent\textbf{Effect of the glyph rendering reference.}
We train a variant without the glyph rendering reference, in which the editing region is erased with a constant value and the target text is provided only as the text condition. As shown in Table~\ref{tab:abl_guidance}, the reference improves Sen.ACC from 0.9013 to 0.9408 and NED from 0.9658 to 0.9862, showing that explicit glyph guidance helps render correct characters.

\begin{table}[t]
\centering
\caption{Ablation on the glyph rendering reference. The reference improves text accuracy.}
\label{tab:abl_guidance}
\small
\setlength{\tabcolsep}{16pt}
\begin{tabular}{l cc}
\toprule
Variant & Sen.ACC$\uparrow$ & NED$\uparrow$ \\
\midrule
w/o reference        & 0.9013 & 0.9658 \\
Ours                 & \textbf{0.9408} & \textbf{0.9862} \\
\bottomrule
\end{tabular}
\end{table}

\noindent\textbf{Effect of depth-normalized recognizer feature supervision.}
We compare our supervision with three variants: no supervision, supervision with a single depth of the recognizer, and supervision with multi-depth VGG features normalized in the same way. As shown in Table~\ref{tab:abl_alignment}, supervising with any single depth improves text accuracy over no supervision, and supervising with all three depths further improves Sen.ACC from 0.9282, obtained by the best single depth, to 0.9408. VGG features also improve over no supervision, but less than the multi-depth features of the text recognizer.

\begin{table}[t]
\centering
\caption{Ablation on the features used for supervision. VGG uses features from multiple depths with the same normalization. Multi-depth text recognizer features achieve the best text accuracy.}
\label{tab:abl_alignment}
\setlength{\tabcolsep}{7pt}
\begin{tabular}{l cc cccc}
\toprule
\multirow{2}{*}{Metric} & \multirow{2}{*}{\begin{tabular}[c]{@{}c@{}}w/o\\supervision\end{tabular}} & \multirow{2}{*}{VGG} & \multicolumn{4}{c}{Recognizer} \\
\cmidrule(lr){4-7}
 & & & Shallow & Middle & Deep & Multi (Ours) \\
\midrule
Sen.ACC$\uparrow$ & 0.9172 & 0.9238 & 0.9245 & 0.9282 & 0.9241 & \textbf{0.9408} \\
NED$\uparrow$     & 0.9685 & 0.9798 & 0.9810 & 0.9815 & 0.9804 & \textbf{0.9862} \\
\bottomrule
\end{tabular}
\end{table}

\noindent\textbf{Effect of depth-wise normalization.}
We train a variant that directly sums the errors of the three depths without normalizing each depth by its ground-truth feature magnitude. As shown in Table~\ref{tab:abl_norm}, this variant obtains a Sen.ACC of 0.9270, within the range of the single depths in Table~\ref{tab:abl_alignment} (0.9241 to 0.9282). Without normalization, combining the three depths thus brings no clear gain over a single depth, consistent with one depth dominating the sum (Figure~\ref{fig:featdist}). With normalization, the combination improves Sen.ACC to 0.9408.

\begin{table}[ht]
\centering
\caption{Ablation on depth-wise normalization in the recognizer feature supervision. Normalizing each depth by its ground-truth feature magnitude improves text accuracy.}
\label{tab:abl_norm}
\small
\setlength{\tabcolsep}{16pt}
\begin{tabular}{l cc}
\toprule
Variant & Sen.ACC$\uparrow$ & NED$\uparrow$ \\
\midrule
w/o normalization    & 0.9270 & 0.9819 \\
Normalized (Ours)    & \textbf{0.9408} & \textbf{0.9862} \\
\bottomrule
\end{tabular}
\end{table}

\section{Conclusion}
\label{sec:conclusion}
In this paper, we propose a video text editing method that formulates the task as text-conditioned video inpainting along a text trajectory. We introduce a trajectory-aligned glyph rendering reference, which provides per-frame glyphs following the position and perspective of the text, and a depth-normalized recognizer feature supervision, which supervises the generated text on multi-depth features of a frozen text recognizer. We further build VTEdit, a real-scene benchmark for video text editing covering text replacement and text addition. Experiments on VTEdit show that our method achieves the best text accuracy and background preservation among image text editing methods, video editing methods, and commercial models, and receives the highest user preference.

\bibliography{iclr2027_conference}

@article{kong2024hunyuanvideo,
  title={{HunyuanVideo}: A systematic framework for large video generative models},
  author={Kong, Weijie and Tian, Qi and Zhang, Zijian and Min, Rox and Dai, Zuozhuo and Zhou, Jin and Xiong, Jiangfeng and Li, Xin and Wu, Bo and Zhang, Jianwei and others},
  journal={arXiv preprint arXiv:2412.03603},
  year={2024}
}

@article{wanteam2025wan,
  title={{Wan}: Open and advanced large-scale video generative models},
  author={{Wan Team} and Wang, Ang and Ai, Baole and Wen, Bin and Mao, Chaojie and Xie, Chen-Wei and Chen, Di and Yu, Feiwu and Zhao, Haiming and Yang, Jianxiao and others},
  journal={arXiv preprint arXiv:2503.20314},
  year={2025}
}

@inproceedings{chen2023textdiffuser,
 author = {Chen, Jingye and Huang, Yupan and Lv, Tengchao and Cui, Lei and Chen, Qifeng and Wei, Furu},
 booktitle = {Advances in Neural Information Processing Systems},
 pages = {9353--9387},
 title = {TextDiffuser: Diffusion Models as Text Painters},
 volume = {36},
 year = {2023}
}

@inproceedings{chen2023diffute,
 author = {Haoxing Chen and Zhuoer Xu and Zhangxuan Gu and Jun Lan and Xing Zheng and Yaohui Li and Changhua Meng and Huijia Zhu and Weiqiang Wang},
 booktitle = {Advances in Neural Information Processing Systems},
 pages = {63062--63074},
 title = {DiffUTE: Universal Text Editing Diffusion Model},
 volume = {36},
 year = {2023}
}

@article{ma2023glyphdraw,
  title={{GlyphDraw}: Seamlessly rendering text with intricate spatial structures in text-to-image generation},
  author={Ma, Jian and Zhao, Mingjun and Chen, Chen and Wang, Ruichen and Niu, Di and Lu, Haonan and Lin, Xiaodong},
  journal={arXiv preprint arXiv:2303.17870},
  year={2023}
}

@article{tuo2024anytext2,
  title={{AnyText2}: Visual text generation and editing with customizable attributes},
  author={Tuo, Yuxiang and Geng, Yifeng and Bo, Liefeng},
  journal={arXiv preprint arXiv:2411.15245},
  year={2024}
}

@article{lan2025fluxtext,
  title={{FLUX-Text}: A simple and advanced diffusion transformer baseline for scene text editing},
  author={Lan, Rui and Bai, Yancheng and Duan, Xu and Li, Mingxing and Jin, Dongyang and Xu, Ryan and Nie, Dong and Sun, Lei and Chu, Xiangxiang},
  journal={arXiv preprint arXiv:2505.03329},
  year={2025}
}

@inproceedings{zhao2025utdesign,
  author       = {Yiming Zhao and
                  Yuanpeng Gao and
                  Yuxuan Luo and
                  Jiwei Duan and
                  Shisong Lin and
                  Longfei Xiong and
                  Zhouhui Lian},
  title        = {{UTDesign}: A Unified Framework for Stylized Text Editing and Generation
                  in Graphic Design Images},
  booktitle    = {Proceedings of the {SIGGRAPH} Asia 2025 Conference Papers},
  pages        = {93:1--93:11},
  year         = {2025}
}

@inproceedings{
geyer2024tokenflow,
title={TokenFlow: Consistent Diffusion Features for Consistent Video Editing},
author={Michal Geyer and Omer Bar-Tal and Shai Bagon and Tali Dekel},
booktitle={International Conference on Learning Representations},
year={2024}
}

@inProceedings{wu2023tuneavideo,
    author    = {Wu, Jay Zhangjie and Ge, Yixiao and Wang, Xintao and Lei, Stan Weixian and Gu, Yuchao and Shi, Yufei and Hsu, Wynne and Shan, Ying and Qie, Xiaohu and Shou, Mike Zheng},
    title     = {Tune-A-Video: One-Shot Tuning of Image Diffusion Models for Text-to-Video Generation},
    booktitle = {Proceedings of the IEEE/CVF International Conference on Computer Vision (ICCV)},
    year      = {2023},
    pages     = {7623--7633}
}

@inProceedings{jiang2025vace,
    author    = {Jiang, Zeyinzi and Han, Zhen and Mao, Chaojie and Zhang, Jingfeng and Pan, Yulin and Liu, Yu},
    title     = {VACE: All-in-One Video Creation and Editing},
    booktitle = {Proceedings of the IEEE/CVF International Conference on Computer Vision (ICCV)},
    year      = {2025},
    pages     = {17191--17202}
}

@inproceedings{bian2025videopainter,
  author    = {Yuxuan Bian and Zhaoyang Zhang and Xuan Ju and Mingdeng Cao and Liangbin Xie and Ying Shan and Qiang Xu},
  title     = {{VideoPainter}: Any-length Video Inpainting and Editing with Plug-and-Play Context Control},
  booktitle = {Proceedings of the ACM SIGGRAPH 2025 Conference Papers},
  pages     = {153:1--153:12},
  year      = {2025}
}

@InProceedings{g2021strive,
    author    = {G, Vijay Kumar B and Subramanian, Jeyasri and Chordia, Varnith and Bart, Eugene and Fang, Shaobo and Guan, Kelly and Bala, Raja},
    title     = {{STRIVE}: Scene Text Replacement in Videos},
    booktitle = {Proceedings of the IEEE/CVF International Conference on Computer Vision (ICCV)},
    year      = {2021},
    pages     = {14549--14558}
}

@article{zeng2026steervte,
  title={SteerVTE: Seamless Video Text Editing with Style and Glyph Control},
  author={Zeng, Kai and Li, Moran and Wang, Zhengwei and Yu, Yingchen and Lin, Yiheng and An, Ruichuan and Lu, Ming and She, Qi and Zhang, Wentao},
  journal={arXiv preprint arXiv:2606.23254},
  year={2026}
}

@inproceedings{lipman2023flowmatching,
  author       = {Yaron Lipman and
                  Ricky T. Q. Chen and
                  Heli Ben{-}Hamu and
                  Maximilian Nickel and
                  Matthew Le},
  title        = {Flow Matching for Generative Modeling},
  booktitle    = {The Eleventh International Conference on Learning Representations,
                  {ICLR} 2023, Kigali, Rwanda, May 1-5, 2023},
  year         = {2023},
}

@inproceedings{tuo2024anytext,
  author       = {Yuxiang Tuo and
                  Wangmeng Xiang and
                  Jun{-}Yan He and
                  Yifeng Geng and
                  Xuansong Xie},
  title        = {AnyText: Multilingual Visual Text Generation and Editing},
  booktitle    = {International Conference on Learning Representations},
  year         = {2024}
}

@inproceedings{ma2026umtext,
  author       = {Lichen Ma and
                  Xiaolong Fu and
                  Gaojing Zhou and
                  Zipeng Guo and
                  Ting Zhu and
                  Yichun Liu and
                  Yu Shi and
                  Jason Li and
                  Junshi Huang},
  title        = {{UM-Text}: {A} Unified Multimodal Model for Image Understanding and
                  Visual Text Editing},
  booktitle    = {Fortieth {AAAI} Conference on Artificial Intelligence},
  pages        = {7791--7799},
  year         = {2026}
}

@article{liu2026innotext,
  author       = {Haowei Liu and
                  Runze He and
                  Jian Lu and
                  Ao Ma and
                  Run Ling and
                  Ke Cao and
                  Jiasong Feng and
                  Wei Feng and
                  Shuo Lu and
                  Yexing Xu and
                  Yun Wang and
                  Jing Wang and
                  Zhanjie Zhang},
  title        = {{InnoText}: {A} Unified Model for Visual Text Generation and Editing},
  journal = {arXiv preprint arXiv:2607.22101},
  year         = {2026}
}

@inproceedings{
wei2026univideo,
title={UniVideo: Unified Understanding, Generation, and Editing for Videos},
author={Cong Wei and Quande Liu and Zixuan Ye and Qiulin Wang and Xintao Wang and Pengfei Wan and Kun Gai and Wenhu Chen},
booktitle={International Conference on Learning Representations},
year={2026}
}

@article{lin2026kiwiedit,
  author       = {Yiqi Lin and
                  Guoqiang Liang and
                  Ziyun Zeng and
                  Zechen Bai and
                  Yanzhe Chen and
                  Mike Zheng Shou},
  title        = {Kiwi-Edit: Versatile Video Editing via Instruction and Reference Guidance},
  journal={arXiv preprint arXiv:2603.02175},
  year         = {2026}
}

@article{klingteam2025klingomni,
title={{Kling-Omni} technical report},
  author={{Kling Team} and Chen, Jialu and Ci, Yuanzheng and Du, Xiangyu and Feng, Zipeng and Gai, Kun and Guo, Sainan and Han, Feng and He, Jingbin and He, Kang and others},
  journal={arXiv preprint arXiv:2512.16776},
  year={2025}
}

@misc{hunyuanvideo2025,
      title={HunyuanVideo 1.5 Technical Report}, 
      author={Tencent Hunyuan Foundation Model Team},
      year={2025},
      eprint={2511.18870},
      archivePrefix={arXiv},
      primaryClass={cs.CV},
      url={https://arxiv.org/abs/2511.18870}, 
}

@article{liu2024glyph,
  		title={Glyph-byt5: A customized text encoder for accurate visual text rendering},
  		author={Liu, Zeyu and Liang, Weicong and Liang, Zhanhao and Luo, Chong and Li, Ji and Huang, Gao and Yuan, Yuhui},
  		journal={arXiv preprint arXiv:2403.09622},
  		year={2024}
	}

@article{Qwen2.5-VL,
  title={Qwen2.5-VL Technical Report},
  author={Bai, Shuai and Chen, Keqin and Liu, Xuejing and Wang, Jialin and Ge, Wenbin and Song, Sibo and Dang, Kai and Wang, Peng and Wang, Shijie and Tang, Jun and Zhong, Humen and Zhu, Yuanzhi and Yang, Mingkun and Li, Zhaohai and Wan, Jianqiang and Wang, Pengfei and Ding, Wei and Fu, Zheren and Xu, Yiheng and Ye, Jiabo and Zhang, Xi and Xie, Tianbao and Cheng, Zesen and Zhang, Hang and Yang, Zhibo and Xu, Haiyang and Lin, Junyang},
  journal={arXiv preprint arXiv:2502.13923},
  year={2025}
}

@inproceedings{hu2022lora,
  author       = {Edward J. Hu and
                  Yelong Shen and
                  Phillip Wallis and
                  Zeyuan Allen{-}Zhu and
                  Yuanzhi Li and
                  Shean Wang and
                  Lu Wang and
                  Weizhu Chen},
  title        = {LoRA: Low-Rank Adaptation of Large Language Models},
  booktitle    = {The Tenth International Conference on Learning Representations, {ICLR}
                  2022, Virtual Event, April 25-29, 2022},
  publisher    = {OpenReview.net},
  year         = {2022},
  url          = {https://openreview.net/forum?id=nZeVKeeFYf9},
  bibsource    = {dblp computer science bibliography, https://dblp.org}
}

@article{li2022pp,
  title={PP-OCRv3: More Attempts for the Improvement of Ultra Lightweight OCR System},
  author={Li, Chenxia and Liu, Weiwei and Guo, Ruoyu and Yin, Xiaoting and Jiang, Kaitao and Du, Yongkun and Du, Yuning and Zhu, Lingfeng and Lai, Baohua and Hu, Xiaoguang and others},
  journal={arXiv preprint arXiv:2206.03001},
  year={2022}
}

@misc{jordan2024muon,
  author       = {Keller Jordan and Yuchen Jin and Vlado Boza and You Jiacheng and
                  Franz Cesista and Laker Newhouse and Jeremy Bernstein},
  title        = {Muon: An optimizer for hidden layers in neural networks},
  year         = {2024},
  url          = {https://kellerjordan.github.io/posts/muon/}
}

@misc{ppocrv62026,
  title={PP-OCRv6: From 1.5M to 34.5M Parameters, Surpassing Billion-Scale VLMs on OCR Tasks},
  author={Yubo Zhang and Xueqing Wang and Manhui Lin and Yue Zhang and Penglongyi Deng and Ting Sun and Tingquan Gao and Zelun Zhang and Jiaxuan Liu and Changda Zhou and Hongen Liu and Suyin Liang and Cheng Cui and Yi Liu and Dianhai Yu and Yanjun Ma},
  year={2026},
  eprint={2606.13108},
  archivePrefix={arXiv},
  primaryClass={cs.CV},
  url={https://arxiv.org/abs/2606.13108},
}

@InProceedings{E_warp_2018_ECCV,
author = {Lai, Wei-Sheng and Huang, Jia-Bin and Wang, Oliver and Shechtman, Eli and Yumer, Ersin and Yang, Ming-Hsuan},
title = {Learning Blind Video Temporal Consistency},
booktitle = {Proceedings of the European Conference on Computer Vision (ECCV)},
month = {September},
year = {2018}
}

@inproceedings{teed2020raft,
  title={Raft: Recurrent all-pairs field transforms for optical flow},
  author={Teed, Zachary and Deng, Jia},
  booktitle={European conference on computer vision},
  pages={402--419},
  year={2020},
}

@article{seedance2026seedance,
  title={Seedance 2.0: Advancing video generation for world complexity},
  author={{Team Seedance} and Chen, De and Chen, Liyang and Chen, Xin and Chen, Ying and Chen, Zhuo and Chen, Zhuowei and Cheng, Feng and Cheng, Tianheng and Cheng, Yufeng and others},
  journal={arXiv preprint arXiv:2604.14148},
  year={2026}
}

@inproceedings{DBLP:journals/corr/SimonyanZ14a,
author       = {Karen Simonyan and
Andrew Zisserman},
editor       = {Yoshua Bengio and
Yann LeCun},
title        = {Very Deep Convolutional Networks for Large-Scale Image Recognition},
booktitle    = {3rd International Conference on Learning Representations, {ICLR} 2015,
San Diego, CA, USA, May 7-9, 2015, Conference Track Proceedings},
year         = {2015},
url          = {http://arxiv.org/abs/1409.1556},
bibsource    = {dblp computer science bibliography, https://dblp.org} }

@ARTICLE{ssim,
  author={Zhou Wang and Bovik, A.C. and Sheikh, H.R. and Simoncelli, E.P.},
  journal={IEEE Transactions on Image Processing}, 
  title={Image quality assessment: from error visibility to structural similarity}, 
  year={2004},
  volume={13},
  number={4},
  pages={600-612},
  doi={10.1109/TIP.2003.819861}}

@inproceedings{DBLP:conf/wacv/RodriguezVLPR23,
  author       = {Juan A. Rodr{\'{\i}}guez and
                  David V{\'{a}}zquez and
                  Issam H. Laradji and
                  Marco Pedersoli and
                  Pau Rodr{\'{\i}}guez},
  title        = {{OCR-VQGAN:} Taming Text-within-Image Generation},
  booktitle    = {{IEEE/CVF} Winter Conference on Applications of Computer Vision, {WACV}
                  2023, Waikoloa, HI, USA, January 2-7, 2023},
  pages        = {3678--3687},
  publisher    = {{IEEE}},
  year         = {2023},
}

@article{DBLP:journals/corr/abs-2409-17524,
  author       = {Chao Li and
                  Chen Jiang and
                  Xiaolong Liu and
                  Jun Zhao and
                  Guoxin Wang},
  title        = {JoyType: {A} Robust Design for Multilingual Visual Text Creation},
  journal      = {CoRR},
  volume       = {abs/2409.17524},
  year         = {2024},
  url          = {https://doi.org/10.48550/arXiv.2409.17524},
  eprinttype   = {arXiv},
  eprint       = {2409.17524},
}

@article{DBLP:journals/corr/abs-2412-17225,
  author       = {Lichen Ma and
                  Tiezhu Yue and
                  Pei Fu and
                  Yujie Zhong and
                  Kai Zhou and
                  Xiaoming Wei and
                  Jie Hu},
  title        = {CharGen: High Accurate Character-Level Visual Text Generation Model
                  with MultiModal Encoder},
  journal      = {CoRR},
  volume       = {abs/2412.17225},
  year         = {2024},
  url          = {https://doi.org/10.48550/arXiv.2412.17225},
  doi          = {10.48550/ARXIV.2412.17225},
  eprinttype   = {arXiv},
  eprint       = {2412.17225},
}

@inproceedings{DBLP:conf/nips/LiuLLLLWWZO25,
  author       = {Jie Liu and
                  Gongye Liu and
                  Jiajun Liang and
                  Yangguang Li and
                  Jiaheng Liu and
                  Xintao Wang and
                  Pengfei Wan and
                  Di Zhang and
                  Wanli Ouyang},
  title        = {Flow-GRPO: Training Flow Matching Models via Online {RL}},
  booktitle    = {Advances in Neural Information Processing Systems 38: Annual Conference
                  on Neural Information Processing Systems 2025, NeurIPS 2025, San Diego,
                  CA, USA, December 2-7, 2025 / Mexico City, Mexico, November 30 - December
                  5, 2025},
  year         = {2025},
}

@inproceedings{DBLP:conf/eccv/ZeilerF14,
  author       = {Matthew D. Zeiler and
                  Rob Fergus},
  title        = {Visualizing and Understanding Convolutional Networks},
  booktitle    = {Computer Vision - {ECCV} 2014 - 13th European Conference, Zurich,
                  Switzerland, September 6-12, 2014, Proceedings, Part {I}},
  volume       = {8689},
  pages        = {818--833},
  year         = {2014},
}

@article{DBLP:journals/corr/abs-2601-03267,
  author       = {OpenAI},
  title        = {OpenAI {GPT-5} System Card},
  journal      = {CoRR},
  volume       = {abs/2601.03267},
  year         = {2026},
  url          = {https://doi.org/10.48550/arXiv.2601.03267},
}
\bibliographystyle{iclr2027_conference}

\clearpage
\appendix
\section{Additional Qualitative Comparisons}
\label{app:qualitative}
We provide additional qualitative comparisons with state-of-the-art methods on text accuracy (Figure~\ref{fig:text_acc}), temporal consistency (Figure~\ref{fig:temp_consistency}), and background preservation (Figure~\ref{fig:bg_preservation}). Our method renders the correct target text, keeps it consistent across frames, and preserves the content outside the editing region, whereas each compared method falls short in at least one of these aspects.

\subsection{Comparison on Text Accuracy}
\label{app:text_accuracy}
\begin{figure}[h]
    \centering

    \includegraphics[width=\linewidth]{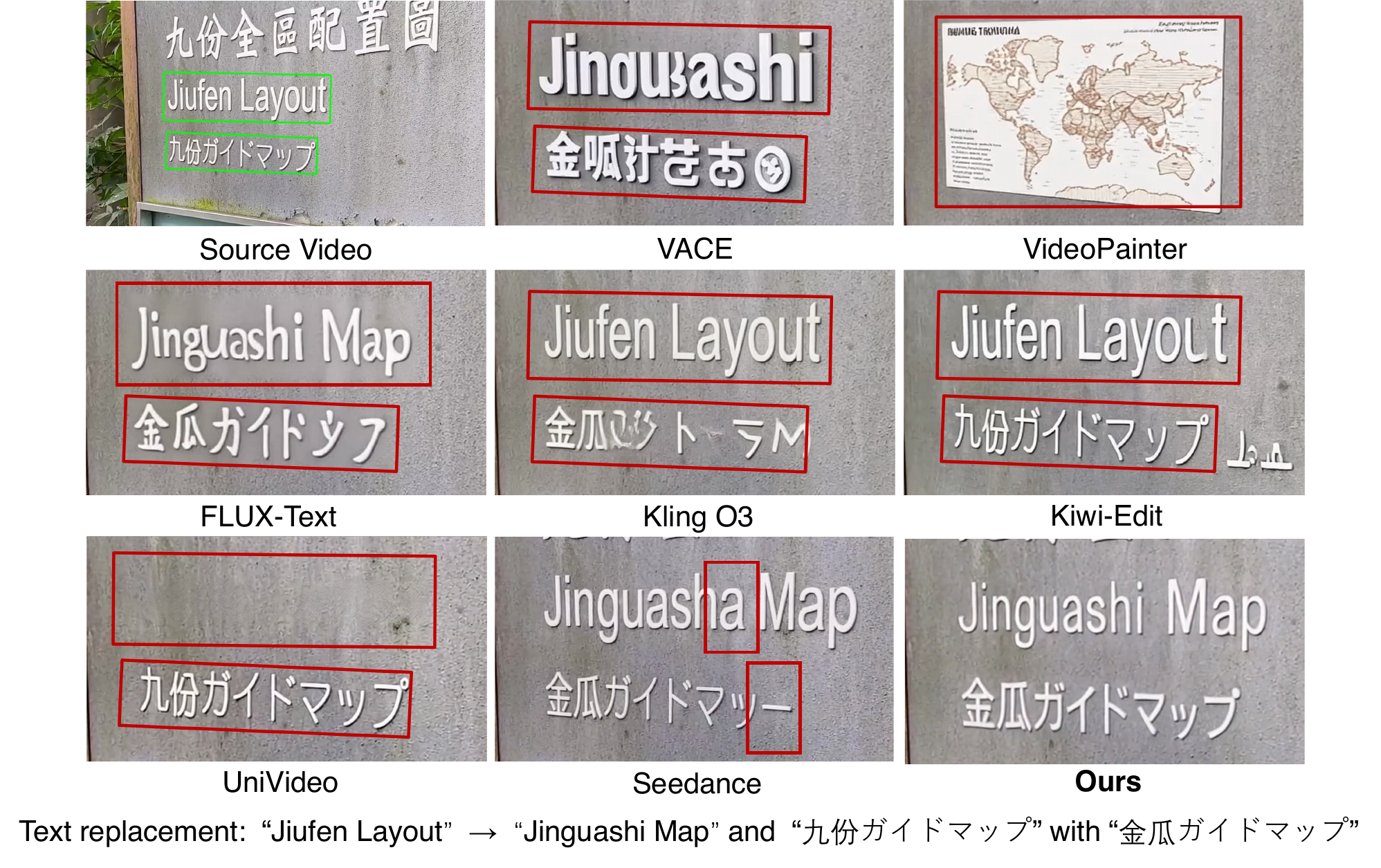}

    \par\vspace{-5pt}
    
    \caption{\textbf{Qualitative comparison of text accuracy.}  \textcolor{green!50!black}{Green boxes} mark the editing regions in the source frame, and \textcolor{red}{Red boxes} highlight unsuccessful edits and character errors. Our method can produce clear and accurate replacement text.}
    \label{fig:text_acc}
\end{figure}

Figure~\ref{fig:text_acc} presents qualitative comparisons with state-of-the-art methods in terms of text accuracy. Seedance produces legible text, but its content does not match the target text. FLUX-Text also generates readable text but suffers from distorted glyphs. VACE, Kling O3, and Kiwi-Edit produce incorrect characters. UniVideo removes the original text rather than completing the requested replacement, while VideoPainter replaces
the lettering with a map graphic. In contrast, our method accurately renders the target text with clear and well-formed glyphs.

\subsection{Comparison on Temporal Consistency}
\label{app:temporal_consistency}

\begin{figure}[h]
    \centering

    \includegraphics[width=\linewidth
    ]{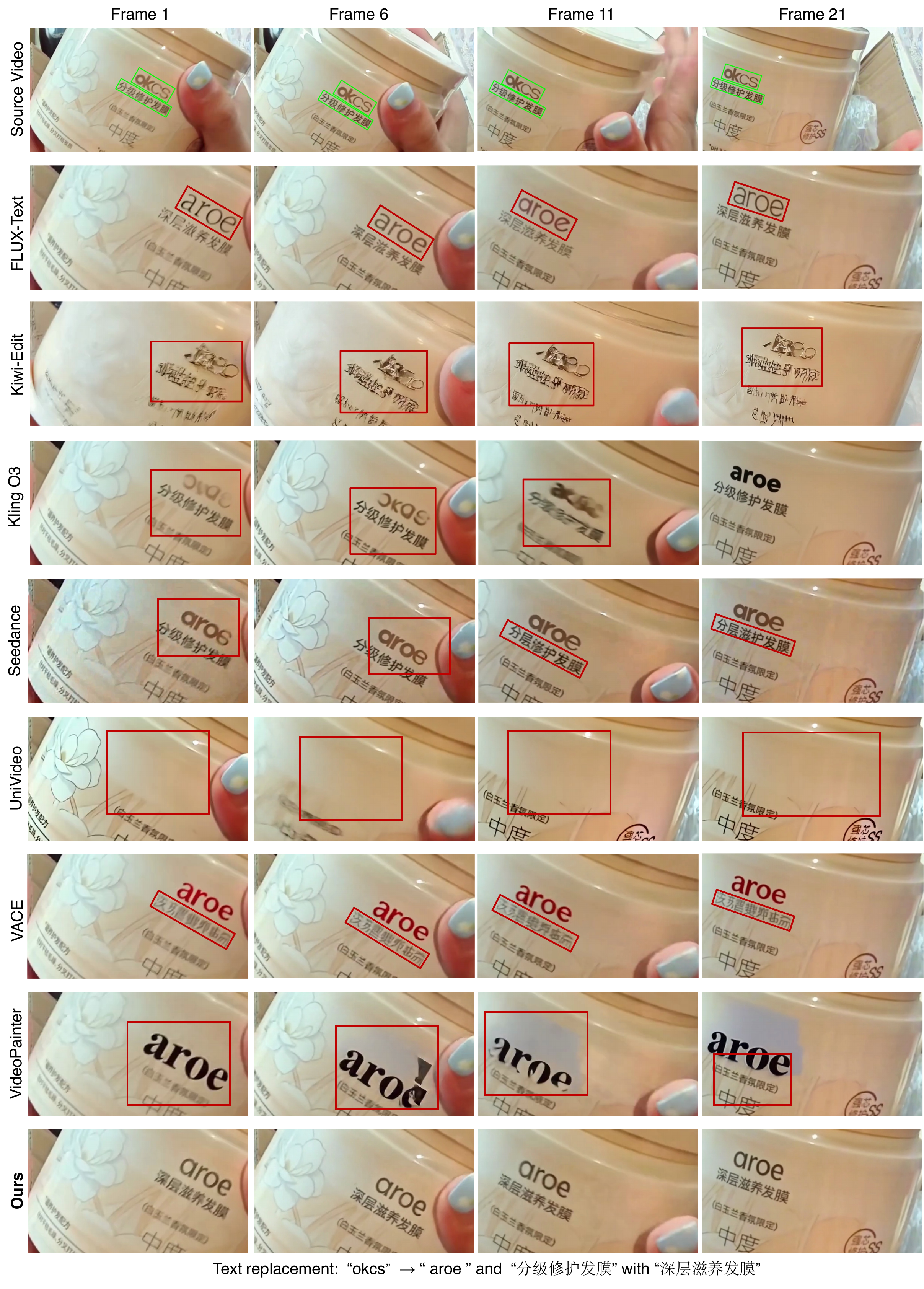}

    \par\vspace{-5pt}
    
    \caption{\textbf{Qualitative comparison of temporal consistency.} Columns show frames 1, 6, 11, and 21. \textcolor{green!50!black}{Green boxes} mark the source editing regions, and \textcolor{red}{Red boxes} highlight failed edits and inconsistent text appearance. Our method maintains legible lettering and a consistent layout across the frames.}
    \label{fig:temp_consistency}
    \vspace{-10pt}
\end{figure}

Figure~\ref{fig:temp_consistency} presents qualitative comparisons with state-of-the-art methods on temporal consistency. Although FLUX-Text renders clear and accurate text in individual frames, its lettering style is inconsistent across frames, resulting in pronounced temporal flickering within the edited regions. Seedance renders the target text correctly in some frames but exhibits glyph errors in others. Other video editing methods either fail to replace the original text or produce results with incorrect characters or noticeable blurring. In contrast, our method accurately renders the target text while maintaining a consistent lettering style across frames, resulting in more temporally coherent edits.

\subsection{Comparison on Background Preservation}
\label{app:background_preservation}

\begin{figure}[t]
    \centering

    \includegraphics[width=\linewidth]{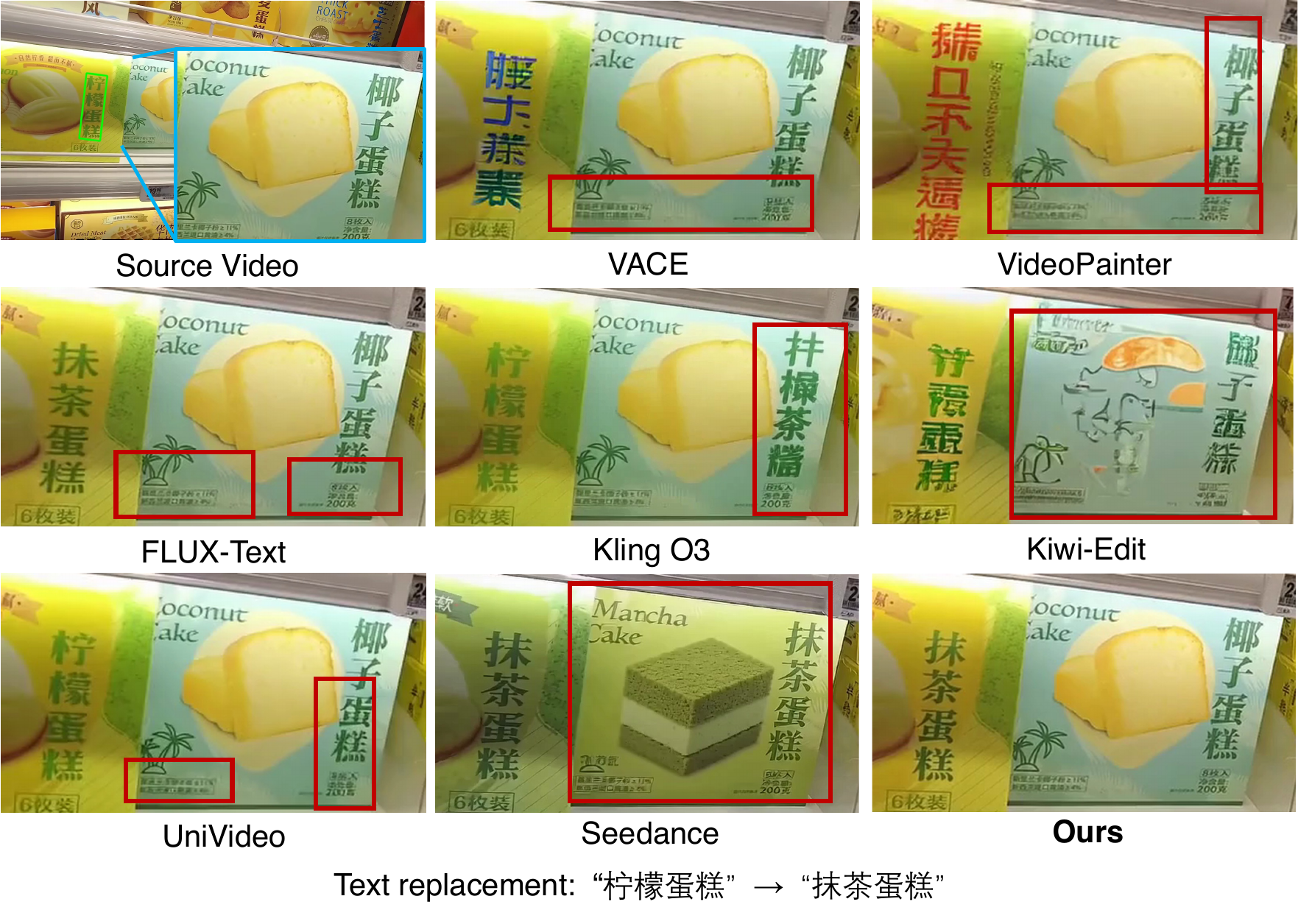}

    \par\vspace{-5pt}
    
    \caption{\textbf{Qualitative comparison of background preservation.} \textcolor{green!50!black}{Green boxes} mark the text to be edited. The \textcolor{blue}{blue inset} enlarges the background outside the editing region. \textcolor{red}{Red boxes} highlight unintended changes to the background. Our method more faithfully preserves the source content outside the editing region.}
    \label{fig:bg_preservation}
\end{figure}

Figure~\ref{fig:bg_preservation} presents qualitative comparisons with state-of-the-art methods on background preservation. The competing methods introduce unintended modifications outside the editing region. Seedance substantially alters the neighboring package, changing its product name, illustration, and color scheme. Kiwi-Edit distorts the package illustration, while Kling O3 incorrectly modifies the neighboring product text. VACE, VideoPainter, FLUX-Text, and UniVideo also corrupt the background text, introducing blurred strokes or malformed characters in regions that should remain unchanged. In contrast, our method edits the target region while more faithfully preserving the surrounding text, illustrations, and visual details.

\section{Implementation Details of Compared Methods}
\label{app:baselines}
FLUX-Text and the mask-based methods take the text boxes as the editing region and the target text in the prompt. The instruction-guided methods take the full source video and an instruction specifying the edit: for text replacement, the instruction gives both the original and the target text, which locates the edit; for text addition, it describes the location of the new text in words, and Kling O3 and Seedance, which accept reference images, are further given a reference frame with the editing region marked by a box. Each compared method is run with its default resolution and number of frames, and its output is aligned to the frames of the source video for evaluation.

\end{document}